%% file: main.tex
\documentclass[acmtog,anonymous=false,review=false]{acmart}
\acmSubmissionID{1234}

\usepackage{booktabs} 

\input{preamble}

\usepackage{graphicx}
\usepackage{booktabs}
\usepackage{enumitem}

\usepackage{soul}
\usepackage{multirow}
\usepackage{colortbl}

\setcopyright{none}

\setcopyright{none}
\renewcommand\footnotetextcopyrightpermission[1]{}

\makeatletter
\AtBeginDocument{%
  \fancypagestyle{firstpagestyle}{%
    \fancyhf{}
    \fancyfoot[C]{\thepage}

  }%
  \fancypagestyle{standardpagestyle}{%
    \fancyhf{}
    \fancyfoot[C]{\thepage}

  }%
  \pagestyle{standardpagestyle}
}
\makeatother

\usepackage[ruled]{algorithm2e} 

\SetAlFnt{\small}
\SetAlCapFnt{\small}
\SetAlCapNameFnt{\small}
\SetAlCapHSkip{0pt}

\acmJournal{TOG}

\begin{document}  
\title{\modelname: 3D Diffusion for Full-Body Avatars from In-the-Wild Videos}

\author{Yiqian Wu}
\authornote{Work was done during an internship at Meta}

\affiliation{%
 \institution{ State Key Laboratory of CAD\&CG, Zhejiang University}  
     \country{China}
 }
 \affiliation{
  \institution{Codec Avatars Lab, Meta} 
  \country{USA}
} 
\email{yiqian.wu.1k@gmail.com}

\author{Rawal Khirodkar} 
 \affiliation{
  \institution{Codec Avatars Lab, Meta} 
  \country{USA}
} 
\email{rawalkhirodkar@gmail.com}

\author{Egor Zakharov} 
 \affiliation{
  \institution{Codec Avatars Lab, Meta} 
  \country{USA}
} 
\email{eozakharov@gmail.com}

\author{Timur Bagautdinov} 
 \affiliation{
  \institution{Codec Avatars Lab, Meta} 
  \country{USA}
} 
\email{timurb@meta.com}

\author{Lei Xiao} 
 \affiliation{
  \institution{Codec Avatars Lab, Meta} 
  \country{USA}
} 
\email{leixiao08@gmail.com}

\author{Zhaoen Su} 
 \affiliation{
  \institution{Codec Avatars Lab, Meta} 
  \country{USA}
} 
\email{suzhaoen@gmail.com}

\author{Shunsuke Saito} 
 \affiliation{
  \institution{Codec Avatars Lab, Meta} 
  \country{USA}
} 
\email{shunsuke.saito16@gmail.com}

\author{Xiaogang Jin} 
\affiliation{%
     \institution{State Key Lab of CAD\&CG, Zhejiang University} 
     \country{China}
     } 
\email{jin@cad.zju.edu.cn}

\author{Junxuan Li}
 \affiliation{
  \institution{Codec Avatars Lab, Meta} 
  \country{USA}
} 
\email{junxuanli@meta.com}

\input{sec/0_abstract}

\keywords{Digital human, 3D diffusion model, Generative 3D human}

\begin{teaserfigure}
  \centering
  \includegraphics[width=0.7\linewidth]{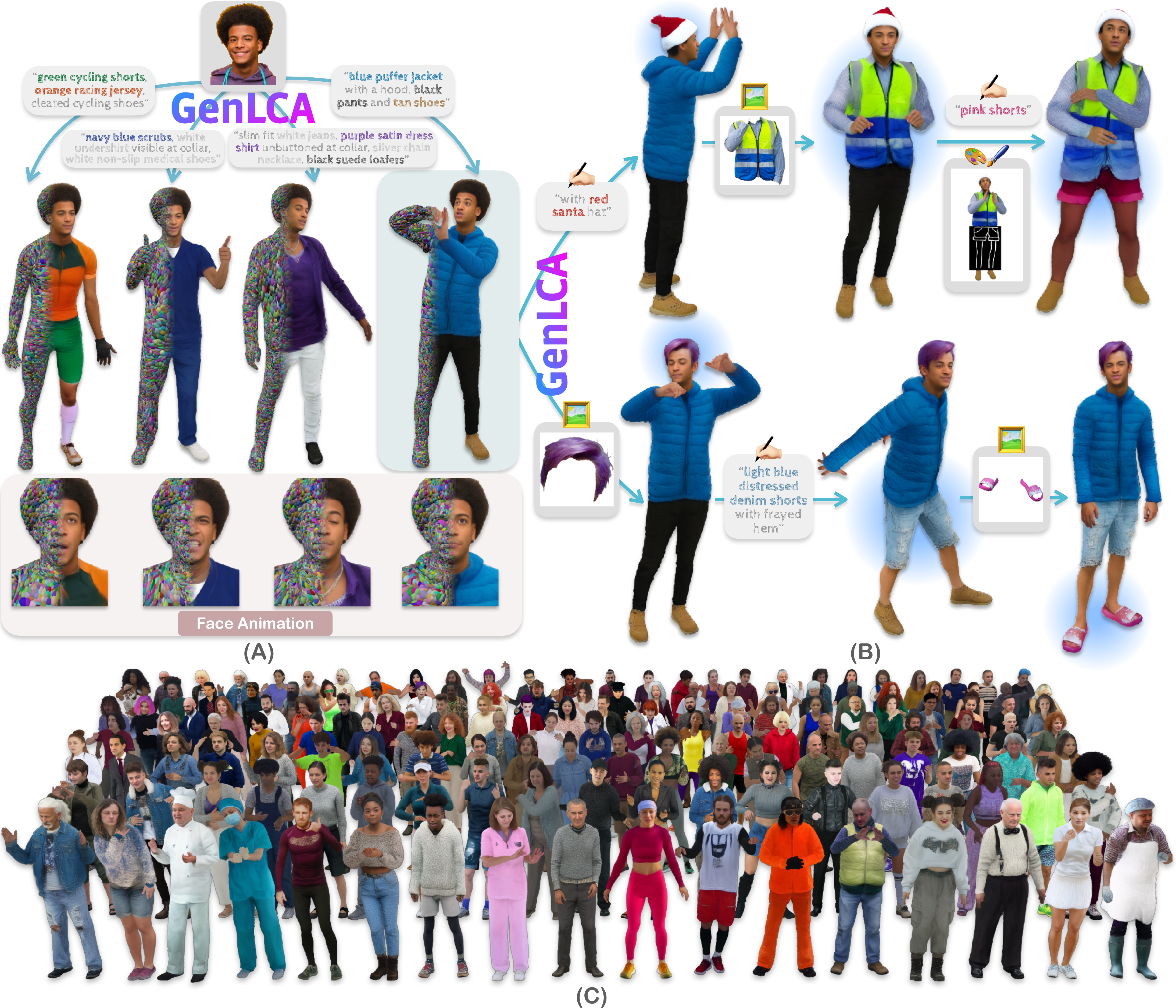}
  \caption{
    \textbf{\modelname} is a diffusion-based generative model for generating and editing full-body 3D Gaussian avatars from text and image inputs.
          (A) \textbf{Generation.} \modelname generates avatars that are visually realistic and consistent with both the identity in the input face image and the semantic descriptions in the input text, while supporting high-fidelity facial and full-body animations. We present zoomed-in and animated face results.
          (B) \textbf{Editing.} By leveraging text, RGB images, or scribbles as control signals, \modelname enables seamless multi-modal editing of the generated avatars.  
          (C) Diverse 3D avatars generated by \modelname from text inputs.  
  }
\label{fig:teaser}
\vspace{-8pt}
\end{teaserfigure}

\maketitle 

\input{sec/1_intro}

\input{sec/2_related_work}

\input{sec/3_methodology}

\input{sec/4_results}
\input{sec/5_conclusion}

\bibliographystyle{ACM-Reference-Format}
\bibliography{main} 

\end{document}

%% file: preamble.tex
\definecolor{applegreen}{rgb}{0.55, 0.71, 0.0}
\definecolor{autumnorange}{rgb}{0.87, 0.61, 0.33}
\definecolor{teal}{rgb}{0.04, 0.73, 0.71} 
\definecolor{2nd}{HTML}{FFECC0}
\definecolor{1st}{HTML}{FFC7A7}

\newcommand{\modelname}{GenLCA\xspace}



%% file: sec/0_abstract.tex
\begin{abstract} 

We present \textbf{\modelname}, a diffusion-based generative model for generating and editing photorealistic full-body avatars from text and image inputs. The generated avatars are faithful to the inputs, while supporting high-fidelity facial and full-body animations. 
The core idea is a novel paradigm that enables training a full-body 3D diffusion model from partially observable 2D data, allowing the training dataset to scale to millions of real-world videos. This scalability contributes to the superior photorealism and generalizability of \modelname. 
Specifically, we scale up the dataset by repurposing a pretrained feed-forward avatar reconstruction model as an animatable 3D tokenizer, which encodes unstructured video frames into structured 3D tokens. 
However, most real-world videos only provide partial observations of body parts, resulting in excessive blurring or transparency artifacts in the 3D tokens.  
To address this, we propose a novel visibility-aware diffusion training strategy that replaces invalid regions with learnable tokens and computes losses only over valid regions. 
We then train a flow-based diffusion model on the token dataset, inherently maintaining the photorealism and animatability provided by the pretrained avatar reconstruction model. 
Our approach effectively enables the use of large-scale real-world video data to train a diffusion model natively in 3D.   
We demonstrate the efficacy of our method through diverse and high-fidelity generation and editing results, outperforming existing solutions by a large margin.
\textbf{The project page is available at \href{https://onethousandwu.com/GenLCA-Page/}{\color{cyan}{GenLCA-Page}}.}

\end{abstract}

%% file: sec/1_intro.tex
\section{Introduction}
\label{sec:intro}
%
As our world becomes increasingly digital, 3D photorealistic avatars hold the key to more natural and expressive virtual experiences. Yet their creation usually requires multi-view images or long monocular videos \cite{Relightable-Full-Body,Animatable-Gaussians, Relightable-Gaussian-Codec-Avatars,IntrinsicAvatar}, which remain inaccessible to most users.
Recent advances in generative models, particularly diffusion models, have shown the ability to create high-quality 3D content from user-friendly inputs, such as text or incomplete images. 
In this paper, our goal is to investigate diffusion models as an efficient and scalable solution for 3D avatar generation.

Diffusion model training benefits from both the high quality and diversity of the training data.
For digital humans, a common solution is to use synthesized data \cite{RODIN, RodinHD, Fake_it_till_you_make_it, IDOL}, but its domain gap from real-world humans often compromises the quality and photorealism of the resulting models \cite{RODIN, RodinHD, TeRA}. To achieve animatability and higher realism, calibrated and synchronized multi-view capture datasets are typically employed \cite{DNA-Rendering, Thuman, Human3.6M}. However, they cover only a few thousand subjects, which impairs model generalization and diversity \cite{SIGMAN, PrimDiffusion, E3Gen}.
As demonstrate in large-scale video diffusion models \cite{Wan-Animate,Hallo3,LTX-Video}, monocular video data provides an ample resource for learning realistic appearance and motion at the 2D level. But training a 3D diffusion model generally requires accurate 3D assets, which remains challenging for monocular videos.


We propose \textbf{Gen}erative \textbf{L}arge 3D \textbf{C}odec \textbf{A}vatar Model (\textbf{\modelname}), a multi-modal 3D diffusion model for generating and editing full-body avatars from text and image inputs, while enabling photorealistic appearance and animation.  
\modelname trains a full-body 3D diffusion model from \textbf{partially observable, million-scale 2D data}, enabled by two key components: a feedforward avatar reconstruction network serving as a \textbf{tokenizer}, and a visibility-aware training strategy to mitigate artifacts in 3D tokens caused by imperfect video frames.

The avatar reconstruction network takes multiple body and face images of a single subject as input. These images are encoded into 3D tokens, which are then decoded by the reconstruction network into an animatable 3D Gaussian avatar. By applying this tokenizer to large-scale video collections, we construct a 3D token dataset for \textbf{$\sim$1.1 million identities}. These tokens are further compressed into compact latents using a compressor to facilitate efficient model training.
However, due to the partial observability of monocular video frames and the reconstruction model’s inherent limitations in hallucinating unobserved regions, the 3D tokens for unobserved areas are often blurry or incomplete. \textit{Directly training a generative model on such imperfect supervision leads to noticeable quality degradation.}
To ensure data quality and fidelity, we introduce a novel visibility-aware training strategy. Specifically, we compute a mask for each identity based on the visibility of their tokens with respect to the input video frames.
Tokens corresponding to unobserved areas are replaced with learnable placeholder features. Furthermore, we apply a masked loss function only to valid regions. This approach limits supervision to observable and reliable 3D regions, thereby mitigating the influence of corrupted information.

We then train a flow-based diffusion model on the compressed latent representations, inheriting the photorealism and animatability of the avatar reconstruction model. To support multi-modal generation and editing, we incorporate three types of modalities as conditional inputs: text, segmented body part images (e.g., hair, face, upper clothing, etc.), and scribble images.

To the best of our knowledge, \modelname is the first method to train a 3D diffusion model at scale using real-world video data. Our method substantially relaxes the data requirements for generative 3D human modeling, \textbf{demonstrating the potential to scale 3D human datasets to a magnitude comparable to that of existing 2D datasets.}
Despite being trained on incomplete observations, \modelname effectively captures semantic relationships between visible tokens, enabling the generation of full-body, animatable avatars, and outperforming SOTAs by a significant margin.

In summary, we make the following contributions:
\begin{itemize}[noitemsep,nolistsep,leftmargin=*]
    \item We propose a 3D diffusion model to produce and edit high-quality, realistic, and animatable 3D full-body avatars from image and text.
    \item We employ a reconstruction model to tokenize unstructured images into 3D tokens, and introduce a novel training strategy that leverages partially observable inputs. Our framework facilitates constructing 3D realistic human datasets at a large scale and paves the way for generalized 3D human generative models. 
\end{itemize}

%% file: sec/2_related_work.tex
\begin{table*}[t]
\caption{\textbf{Comparison of training datasets for 3D human diffusion models.} Our dataset contains the largest number of identities and the most realistic data among existing methods.  ``$^*$'' indicates that the corresponding model is trained on each sub-dataset individually, rather than on a mixture of them.
}
\centering
\scalebox{0.99}{
\begin{tabular}{@{}ccccc@{}}
\toprule
\multicolumn{1}{c}{Dataset}   & \# total IDs  & \# synthetic data   & \# captured data   & \# in-the-wild data         \\ \midrule
StructLDM \cite{StructLDM} & 1.8K  & 0.8K$^*$ &  0.5K$^*$ & 0.5K$^*$\\
HumanLiff \cite{HumanLiff} & 1.1K & 1K$^*$ &  0.1K$^*$ & 0\\
Rodin \cite{RODIN} & 100K & 100K & 0  & 0  \\
RodinHD \cite{RodinHD} &  46K&  46K& 0  & 0  \\
SimAvatar \cite{SimAvatar} & 20K & 20K & 0  & 0  \\
TeRA \cite{TeRA} &  70K  &    70K      & 0  & 0  \\
SIGMAN \cite{SIGMAN}  & 110K  & 100K    &  10K & 0  \\
\modelname (Ours)  &\textbf{1,117K}  & 0   &   4K &  1,113K    \\ \bottomrule
\end{tabular}
}
\label{tab:dataset_compare}
\end{table*}
\section{Related Work}
\label{sec: Related Work}

\subsection{3D human reconstruction} 
Existing work has explored a variety of 3D representations to achieve high fidelity zero-shot avatar reconstruction and re-animation, including parametric meshes \cite{nphm, smplx,flame}, neural radiance fields (NeRF) \cite{RigNeRF, KeypointNeRF}, and 3D Gaussian splatting (3DGS) \cite{GaussianAvatars, SplattingAvatar,Relightable-Full-Body,RePerformer,Drivable-3D-Gaussian-Avatars}.
By conditioning these 3D representations on controllable parameters such as pose and lighting, dynamic details can be integrated into the avatar \cite{NerFACE, Gaussian-Head-Avatar,NPGA,Relightable-Full-Body,Animatable-Gaussians}, enabling the creation of more expressive models. 
However, these approaches require extensive camera coverage \cite{Codec_Avatar_Studio, DNA-Rendering, Human3.6M, Thuman}, disentangled attribute supervision, and sufficient capture of fine-grained details to achieve high-quality results.
Feedforward reconstruction models focus on training 3D human priors~\cite{LHM, PF-LHM, IDOL, URAvatar, GAGAvatar,HumanRAM} to directly regress 3D human representations from 2D inputs in a single forward pass.
\textbf{However, these reconstruction models fail to produce high-quality results for unobserved regions and lack editability.}

In contrast to zero-shot or feed-forward reconstruction methods, \modelname requires minimal input during inference and supports both generation and editing.

\subsection{Zero-shot and one-shot 3D human creation}
DreamFusion \cite{DreamFusion} introduces Score Distillation Sampling (SDS) for generating 3D content using guidance from 2D diffusion models \cite{ddpm,latent-diffusion-model}.
SDS-based 3D human creation utilizes 3D representations such as meshes \cite{TADA, HumanNorm, Tech}, neural radiance fields \cite{portrait3d, AvatarVerse}, and 3D Gaussian splatting (3DGS) \cite{HumanGaussian, DreamWaltz-G, HeadStudio,avatargo}, followed by multi-step SDS optimization. 
However, SDS optimizes a single 3D content using a 2D diffusion prior, leading to ambiguities that cause over-saturation and unrealistic styles.
Another line of research aims to directly reconstruct 2D avatars from multi-view data hallucinated by 2D diffusion models \cite{CAP4D, Joker, PERSE, PSHuman, Human-3Diffusion, AdaHuman,FaceLift} or video diffusion models \cite{Zero-1-to-A, Diffuman4D, GAS}. 
To ensure geometric alignment and reduce view inconsistency, these methods either condition the diffusion model on 3D control signals \cite{Joker, CAP4D, PERSE, Pippo,Diffuman4D} or incorporate reconstruction into the denoising process \cite{AdaHuman, Human-3Diffusion, Gen-3Diffusion}. Nevertheless, inherent view inconsistencies still result in blurriness in the final outputs.

Compared to the aforementioned generation methods that rely on 2D diffusion models, \textbf{our \modelname operates natively in 3D and is trained on real-world data, thereby inherently avoiding issues of blurriness and low realism}.

\subsection{Generative 3D human model}
Inspired by EG3D \cite{eg3d}, which employs GANs \cite{GANs} to generate implicit neural fields and uses 2D images for supervision, numerous works \cite{AniPortraitGAN,AG3D,3DHumanGAN,EVA3D,En3D,Gaussian-Shell-Maps,XAGen} have extended it to full-body human modeling. However, directly modeling 3D implicit distribution from single-view 2D collections introduces ambiguities and leads to quality degradation. 
Diffusion models \cite{ddpm,latent-diffusion-model} have been extended to 3D human modeling \cite{PrimDiffusion, StructLDM, TeRA, SIGMAN,RODIN, RodinHD,SimAvatar}. 
Since diffusion model training requires accurate 3D assets for each training sample, a typical training pipeline involves an encoding process. This process either trains an auto-encoder or performs zero-shot optimization to encode multi-view images into structured  representations, such as feature planes \cite{RODIN, RodinHD, HumanLiff}, structured UV latents \cite{StructLDM, TeRA, SIGMAN, MoGA,E3Gen,tang_3d_ani_human_geo_dist,tang_3d_human_geo_dist}, or 3D primitives \cite{PrimDiffusion,GaussianCube}.
However, since their encoding process is designed to accurately represent each identity in existing datasets, these methods require extensive camera coverage to achieve optimal performance and therefore cannot be generalized to in-the-wild data.
Consequently, their training sources are limited to small-scale captured datasets~\cite{StructLDM, SIGMAN,HumanLiff,E3Gen,tang_3d_ani_human_geo_dist,tang_3d_human_geo_dist} or unrealistic synthetic datasets~\cite{RODIN, PrimDiffusion, RodinHD, TeRA, SIGMAN, SimAvatar,HumanLiff}, as shown in Tab. \ref{tab:dataset_compare}.

In summary, there is no 3D avatar generator that performs native 3D generation while effectively utilizing in-the-wild data.  
To address this limitation, we propose to leverage a large-scale reconstruction model to extract training samples from in-the-wild videos. This approach substantially expands the available dataset for a more generalized 3D human diffusion model.

%% file: sec/3_methodology.tex
\section{Methodology}
\label{sec: Methodology} 
%
In this section, we first introduce the 3D avatar tokenizer, which encodes images into 3D tokens (Sec. \ref{sec: recon_model}).
Next, we present the overall architecture and training strategy of \modelname (Sec. \ref{sec: generative_model}). 
\modelname first employs a compressor to compress the 3D tokens into compact latents.
To mitigate the influence of corrupted information in these 3D tokens, we propose a visibility-aware training strategy that utilizes a visibility mask to restrict training to observable and reliable 3D information. 
We then detail the generative model architecture and describe our conditional inputs.
%
Finally, we explain the computation of the visibility mask (Sec. \ref{sec: visibility mask}).

\begin{figure}[t]
  \centering
  \includegraphics[width=0.99\linewidth]{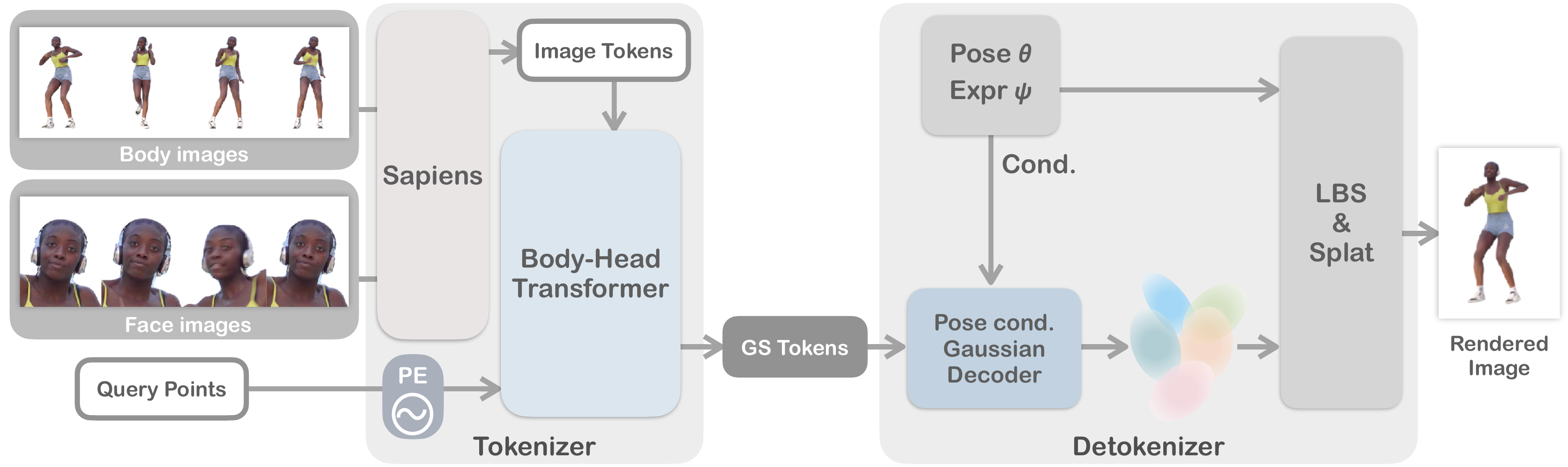}
  
  \caption{ 
   \textbf{The architecture of the reconstruction model.} The transformer takes image tokens and query point embeddings as inputs, and outputs GS tokens. The GS tokens are decoded to get dynamic GS attributes. 
   The resulting Gaussian splats are rendered using LBS to obtain the final renderings.
   }
  \label{fig:tokenizer}
  \vspace{-15pt}
\end{figure}

\subsection{3D avatar tokenizer}
\label{sec: recon_model} 
To obtain a structured and unified 3D avatar representation from 2D images, we propose leveraging a pre-trained reconstruction model, \textbf{LCA} \cite{li2026largescalecodecavatarsunreasonable}, as the \textit{tokenizer}.
LCA is a reconstruction-based model designed to produce high-fidelity 3D tokens for input frames. 
Further details are provided in Appendix B of the supplementary material.

We illustrate the high-level architecture of {LCA} in Fig.~\ref{fig:tokenizer}. The model can be divided into two components: the tokenizer and the detokenizer. The tokenizer encodes multiple input images into 3D tokens, while the detokenizer interprets these tokens as Gaussian splats.
Specifically, the inputs to the tokenizer consist of multiple body and face images. Sapiens~\cite{Sapiens} then extracts image tokens from the input images.  
Subsequently, $N$ query points are sampled on a template body mesh, denoted as $\mathbf{X} = \{ x_i \in \mathbb{R}^{3} \}$, which is fixed and shared across all identities.
Images tokens and point embeddings are fed into a transformer, which produces $N$ GS tokens $\mathbf{T} = \{ t_i \in \mathbb{R}^{D_{\mathbf{T}}} \}$. Each GS token $t_i$ is mapped to its corresponding query point $x_i$.
During detokenization, each GS token is decoded by a lightweight MLP-based decoder into eight Gaussian splats, resulting in a total of $8N$ splats per identity. The decoder is further conditioned on pose and expression parameters to enable dynamic GS features.
Finally, LBS and splatting are used to render the final image.
During tokenization, we use four body images and four face images as input, which are extracted from video data as detailed in Sec. \ref{sec: Dataset}. The number of query points is set as $N=8,192$, while the dimention of token is $D_{\mathbf{T}}=1,024$.

\begin{figure*}[t]
  \centering
  \includegraphics[width=0.99\linewidth]{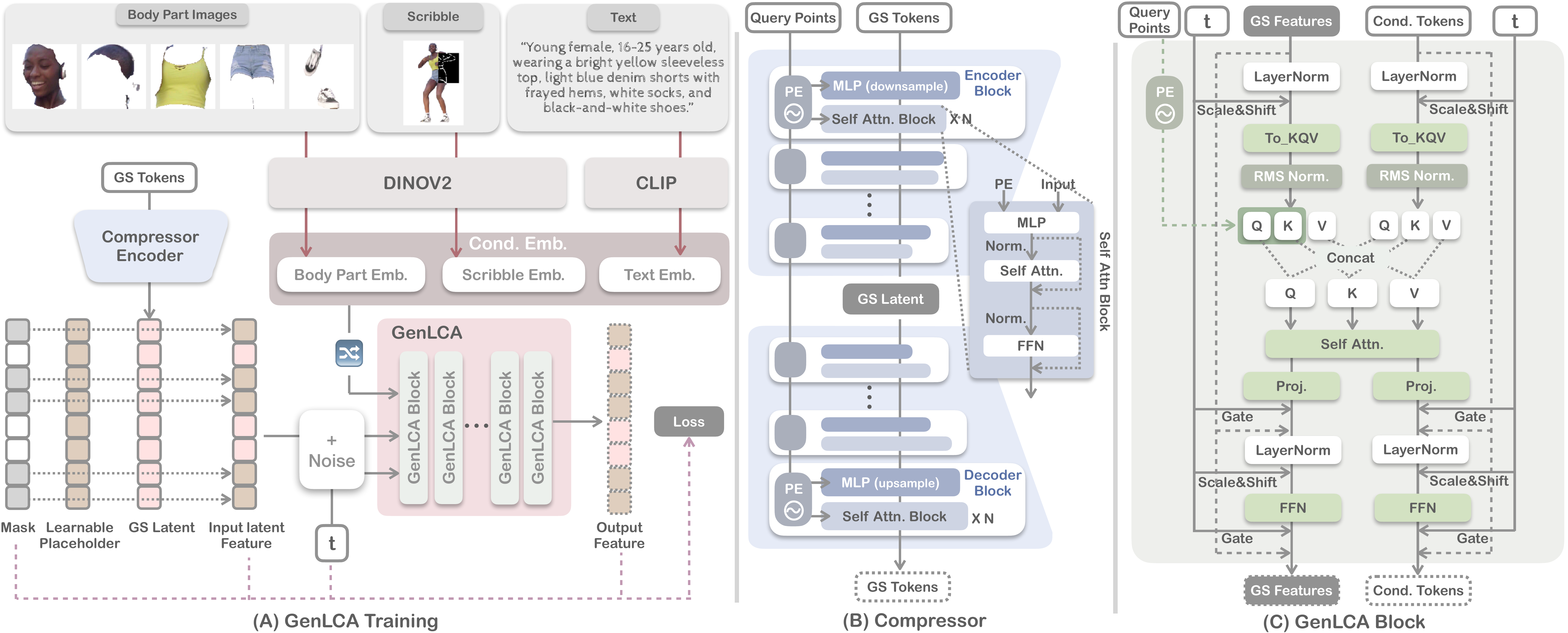}
  
  \caption{ 
    (A) \textbf{Training pipeline of \modelname.} During training, the high dimensional GS tokens are first encoded into compact GS latent by the compressor encoder. For conditional inputs, we use CLIP \cite{clip} to extract text embeddings and DINOv2 \cite{DINOv2} to extract scribble and body part embeddings.
    To prevent the training process from being affected by corrupted information, we replace invalid regions (as indicated by the visibility mask) with learnable placeholder features and employ a masked loss.
    (B) \textbf{Detailed architecture of the compressor.} The compressor’s encoder and decoder consist of MLPs for downsampling or upsampling, combined with self-attention blocks for feature fusion. Positional encoding is applied within each block.
    (C) \textbf{Detailed architecture of the \modelname block.} We adapt the MMDiT block as the basic block of \modelname. Each \modelname block takes the query points, time step, latent features, and conditional features as inputs. Separate branches are used to process latent and conditional features, and positional encoding is added to the latent features.
   }
  \label{fig:network}
\end{figure*}
 
\subsection{\modelname}
\label{sec: generative_model} 
\modelname is a flow-based diffusion model trained with the rectified flow objective \cite{flow_matching}.  
In this section, we detail the training strategy and model architecture of \modelname.

\subsubsection{Token compressor}
\label{sec: compressor}
The extracted GS tokens $\mathbf{T} \in \mathbb{R}^{N \times D_{\mathbf{T}}}$ are high-dimensional representations that are scattered in a loosely structured space. To obtain a more compact space for generative model training, we use a compressor to encode the GS tokens into latents.

The detailed architecture of the compressor is presented in Fig. \ref{fig:network} (B). Both the encoder and decoder are composed of multiple blocks, each containing a MLP for downsampling or upsampling, as well as self-attention blocks for feature fusion. Additionally, the same set of query points $\mathbf{X} \in \mathbb{R}^{N \times 3}$ used during tokenization are encoded to obtain positional embeddings. 
The compressor is trained with the following loss function:
\begin{equation}
\begin{split}
\label{eqn: compressor}
\mathcal{L}_{\text{compressor}} = \lambda_{1}\mathcal{L}_1(\mathcal{D}(\mathcal{E}(\mathbf{T}, \mathbf{X}), \mathbf{X}), \mathbf{T}) + \lambda_{2} \mathcal{L}_{\text{KL}}, 
\end{split} 
\end{equation}
where $\mathcal{D}$ and $\mathcal{E}$ denote the encoder and decoder, respectively. The $\mathcal{L}_1$ reconstruction loss is computed between the reconstructed tokens $\mathcal{D}(\mathcal{E}(\mathbf{T}, \mathbf{X}), \mathbf{X})$ and the ground truth tokens $\mathbf{T}$. The KL divergence loss $\mathcal{L}_{\text{KL}}$ is also included.  $\lambda_{1}$ and $\lambda_{2}$ are the corresponding weights for the losses.  
The compressed latents $\mathbf{Z} = \mathcal{E}(\mathbf{T}, \mathbf{X}) \in \mathbb{R}^{N \times D_{\mathbf{Z}}}$ have the same number of tokens as $\mathbf{T}$, but with a lower dimension $D_{\mathbf{Z}} = 8$.
 
%
%

\subsubsection{Visibility-aware training}
\label{sec: Visibility-aware training}
After training the compressor, \modelname is trained in the latent space.
However, due to the partial observability of monocular video frames and the inherent limitations of the LCA reconstruction model in hallucinating unobserved regions, the information for these areas is often blurry or incomplete, as discussed in Sec. \ref{sec: visibility mask}.
Directly training a generative model on such imperfect supervision leads to noticeable quality degradation (discussed in Sec. \ref{sec: Ablation_studies}).
To ensure data quality and fidelity, we propose a novel visibility-aware training strategy, which utilizes a visibility mask (Sec. \ref{sec: visibility mask}) to apply different training strategies to latents corresponding to observable and unobservable regions.
As shown in Fig. \ref{fig:network} (A), we introduce learnable placeholder features that are shared among all identities. To mitigate the influence of invalid information, we replace invalid latent components with these placeholder features. Additionally, during loss computation, we use masked weighting to ensure that the loss is computed only over valid regions.

\subsubsection{Model architecture}
\label{sec: Model_Architecture}
The detailed architecture of \modelname is illustrated in Fig. \ref{fig:network} (C). We adopt the double-stream MMDiT block \cite{MMDiT} from Hunyuan \cite{hunyuan3d}. In this design, the latent features and conditional tokens are processed by separate network branches (distinct branches are used for different modalities, this is omitted from the figure for clarity) to obtain query, key, and value features. These features are concatenated and used to perform attention.
These attention outputs are split into latent and conditional components, each processed by separate branches to produce block outputs, which are then fed into the next block.

For modulation, we only use the time step. Additionally, we add point embeddings to the query and key features of the latents, as our latent features maintain a one-to-one correspondence with the associated query points.

\subsubsection{Conditional inputs}
\label{sec: Conditional_inputs}

For conditional inputs, as illustrated in Fig. \ref{fig:network} (A), we utilize text descriptions, scribble images, and  body part images. For text, we use CLIP \cite{clip} to extract text embeddings $\mathbf{C}_{\text{text}}$. For scribble images, we use DINOv2 \cite{DINOv2} to extract scribble embeddings $\mathbf{C}_{\text{scribble}}$. Similarly, for body part images, we input five body part images into DINOv2 and concatenate the resulting embeddings to obtain the body part embeddings $\mathbf{C}_{\text{body}}$.

To enable flexible controllability, we design three types of input modalities: \textit{text-only}, \textit{image-only}, and \textit{text-plus-image}. During training, one of these modalities is uniformly sampled as the conditional input.
If the selected modalities involves images (\textit{image-only} or \textit{text-plus-image}), we further uniformly sample between scribble images and body parts as the image input.
%

\begin{figure}[t]
  \centering
  \includegraphics[width=0.99\linewidth]{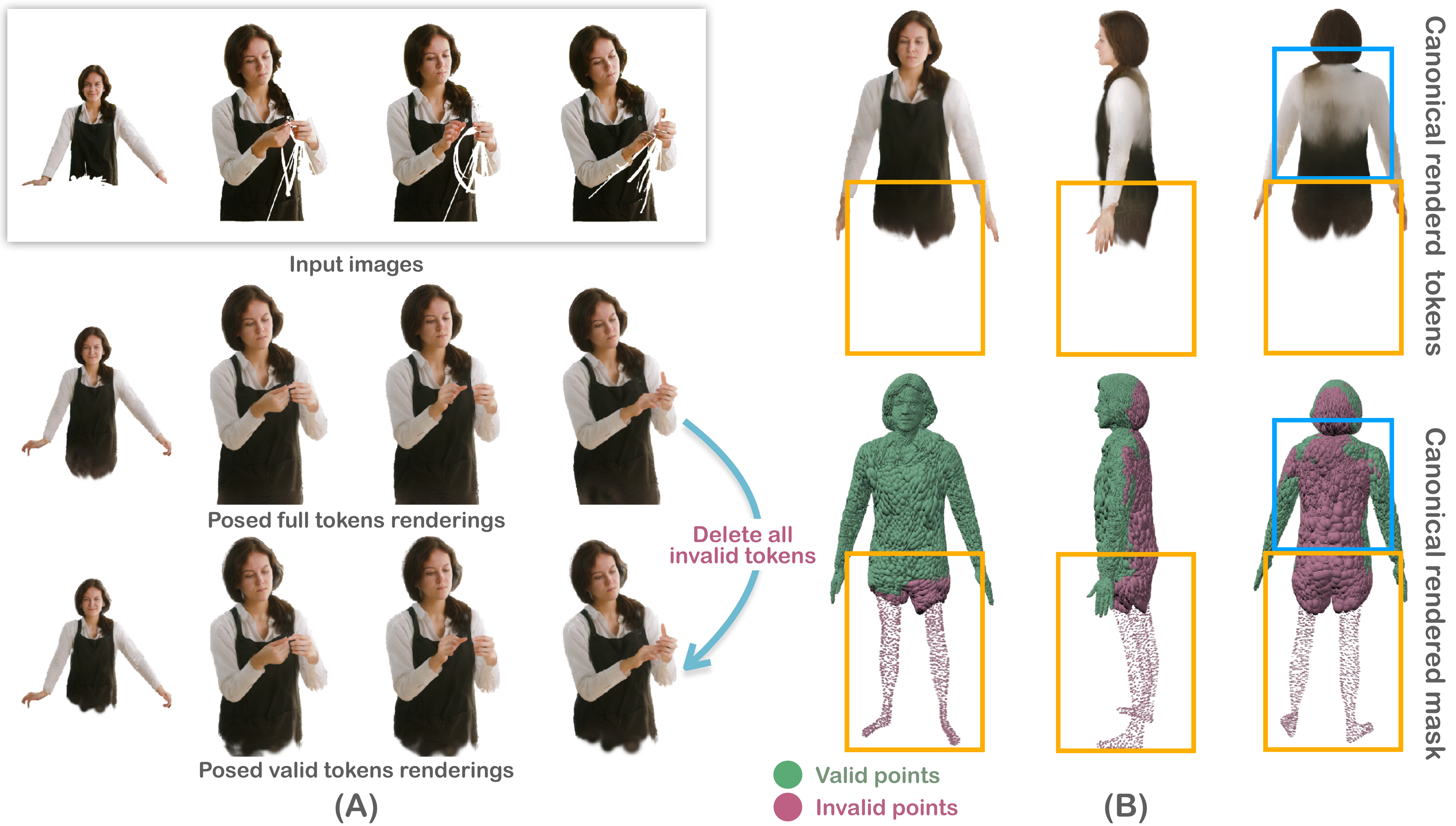}
  
  \caption{ 
  (A) The tokens accurately reconstruct the visible regions of the images. After filtering out all invalid tokens and retaining only the valid ones, the rendered results still achieve high-quality reconstruction. (B) We present the rendered tokens in canonical space (1st row), where blurry regions are highlighted with  {blue boxes} and transparent regions with {yellow boxes}. The visibility mask (2nd row) separates the valid and invalid regions. 
   }
  \label{fig:masking}
\end{figure}

 \begin{figure*}[!t]
  \centering
  \includegraphics[width=0.8\linewidth]{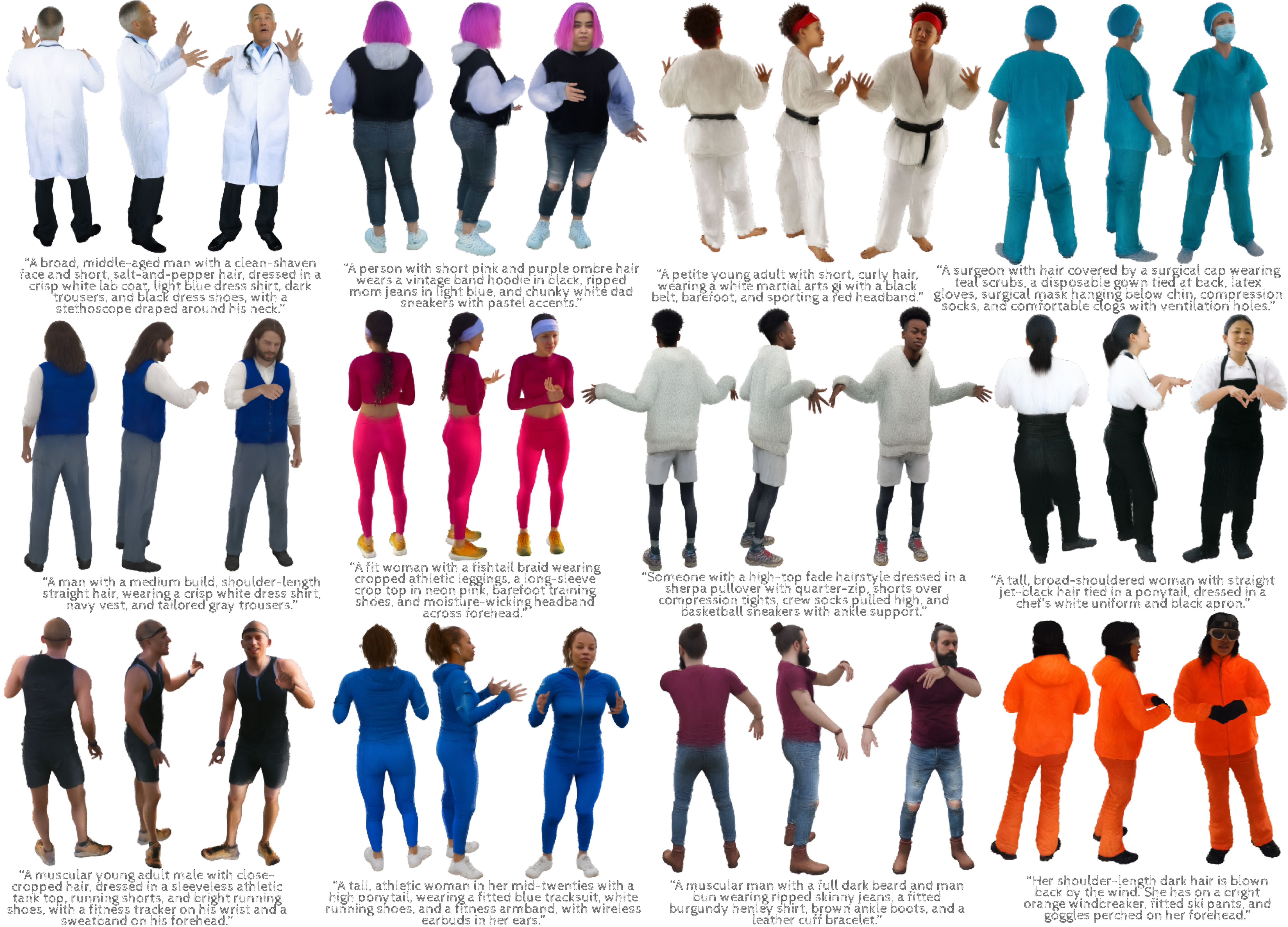}
  
  \caption{ 
   \textbf{3D avatars generated by \modelname from texts.} All results are generated with CFG scale = 5.0, 50 sampling steps, and animated with random poses.  
   }
  \label{fig:results}
\end{figure*}

\subsection{Visibility mask}
\label{sec: visibility mask}

As previously discussed, the tokenizer exhibits limitations in hallucinating unobserved regions. Consequently, artifacts tend to appear in areas with limited image information (Fig. \ref{fig:masking} (B)). As described in Sec. \ref{sec: Visibility-aware training}, we employ a visibility-aware training strategy to address this issue, which requires a visibility mask to label invalid regions. In this section, we provide a detailed explanation of the visibility mask calculation.

As shown in Fig.~\ref{fig:masking} (B), when examining the complete set of tokens, we observe blurry back regions (highlighted by blue boxes) due to the input images being exclusively frontal views, and transparent lower body regions (highlighted by yellow boxes) resulting from the input images containing only upper body views. 
To obtain the visibility mask, we render the decoded Gaussian splats using the corresponding camera and body poses from the input body image and compute the gradients, which indicate each splat's contribution to the rendered image. Splats with low contributions are considered to have low ``visibility'' relative to the input image.
A token is defined as visible if at least two out of eight decoded splats are visible in at least one of the input views. This process produces the visibility mask, as illustrated in the second row of Fig. \ref{fig:masking} (B). The corresponding Gaussian splats of filtered valid tokens are of high quality and appear realistic, whereas the invalid ones are typically blurry or even transparent.

\section{Implementation Details}

\subsection{Training dataset}
\label{sec: Dataset} 
We construct the training dataset for \modelname by encoding frames from monocular videos into structured 3D tokens using the tokenizer. 
Please refer to the supplementary material for visual examples.

Specifically, we reuse the video dataset employed for training LCA \cite{li2026largescalecodecavatarsunreasonable} to construct our training token dataset. The video dataset comprises:
\begin{itemize}[noitemsep,nolistsep,leftmargin=*]
    \item \textbf{In-the-wild data.} A total of 1,113,476 monocular, human-centric real-world videos are included to ensure diversity and broad generalization. 

 \item  \textbf{Captured data.}  
To provide cleaner data with comprehensive full-body coverage, the dataset additionally contains calibrated and synchronized multi-view videos of 2,737 identities recorded in a studio capture setup similar to \cite{Codec_Avatar_Studio}. Furthermore, 1,198 individuals are recorded using mobile phones, where participants perform a full-body rotation to ensure complete coverage from diverse viewpoints.  
\end{itemize}

For each identity, we select frames with the largest differences in yaw angles as body input images, maximizing the coverage of observable body regions. Additionally, we randomly sample multiple frames from the video and crop the face region to serve as face input images. These images are processed by the tokenizer to obtain GS tokens $\mathbf{T}$. The final token dataset comprises \textbf{1,117,411} identities. For evaluation purposes, we sample 1,000 high-quality identities from the captured dataset to serve as our test set.

For each input image, we use Sapiens \cite{Sapiens} for body segmentation and background removal. 
For multi-modal generation and editing, each image is annotated with three types of labels: text descriptions, scribble images, and body part images. 
Please refer to the supplementary material for further annotating details.
 
 \begin{figure*}[t]
  \centering
  \includegraphics[width=0.7\linewidth]{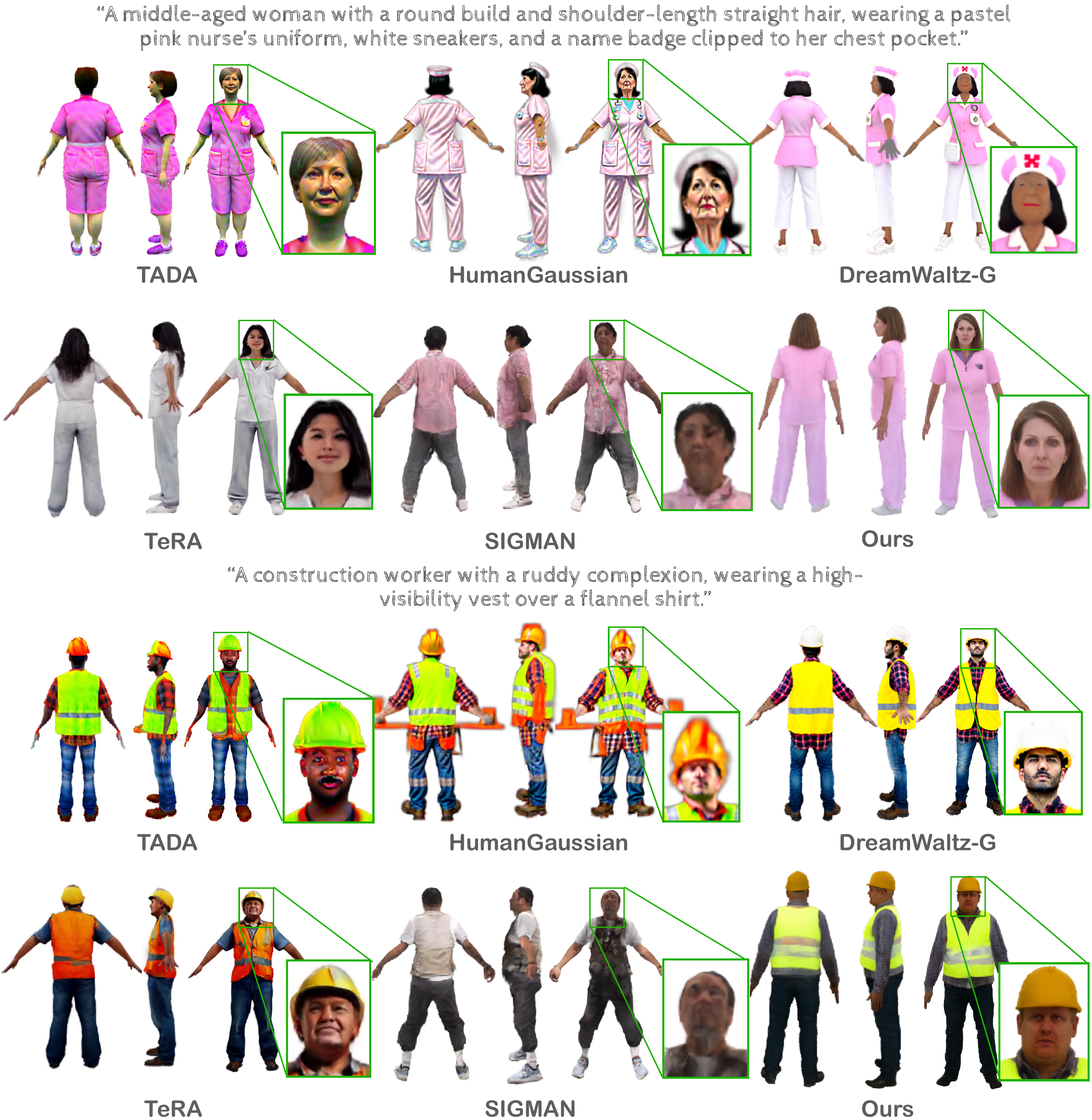}
  
  \caption{ 
  We compare our \modelname with SOTA methods, including SDS based approaches: TADA \cite{TADA}, HumanGaussian \cite{HumanGaussian}, and DreamWaltz-G \cite{DreamWaltz-G}, and text conditioned 3D human diffusion models, TeRA \cite{TeRA} and SIGMAN \cite{SIGMAN}. We use the same text prompt as input. In addition to full-body renderings, we also provide zoomed-in views for comparison of facial regions.  
   }
  \label{fig: comparison}
\end{figure*}

\subsection{Model architecture and training details}
\noindent\textbf{Compressor.} Our compressor maps the token $\mathbf{T} \in \mathbb{R}^{8192 \times 1024}$ to a latent representation $\mathbf{Z} \in \mathbb{R}^{8192 \times 8}$ via an encoder, and reconstructs $\mathbf{T}' \in \mathbb{R}^{8192 \times 1024}$ using a decoder. The encoder consists of seven blocks with progressively reduced channel dimensions 512, 256, 128, 64, 32, 16, 8.
The decoder contains five blocks with channel dimensions 32, 64, 128, 512, 1024. The number of tokens (8,192) remains constant throughout.
We use SiLU activations and Layer Normalization in all MLP layers and at the input of each self-attention block. 
The compressor is trained on 32 NVIDIA A100 GPUs with a batch size of 256 for one day. The learning rate is linearly warmed up from $4 \times 10^{-10}$ to $4 \times 10^{-4}$ over the first 1K iterations. The reconstruction loss weight is set to 1.0, while the KL divergence weight is increased linearly from $1 \times 10^{-3}$ to $1 \times 10^{-2}$ over 10K iterations.

\noindent\textbf{\modelname.} The denoising network of \modelname consists of 28 blocks, each with 1,024 channels, 16 attention heads, and an FFN with an MLP ratio of 4.0. RMSNorm \cite{rms_norm} is applied to the query and key features. The number of latent tokens (8,192) remains constant across all blocks.
For conditional input tokenization, we use the huge version of MetaCLIP and the big version of DINOv2 with registers.
\modelname is trained with the rectified flow objective using the Conditional Flow Matching (CFM) loss \cite{flow_matching} with $\sigma_{\text{min}} = 1 \times 10^{-5}$. Training is performed on 64 NVIDIA A100 GPUs with a batch size of 128 for four days. The learning rate is linearly warmed up from $2 \times 10^{-10}$ to $2 \times 10^{-4}$ over the first 1K iterations.
Classifier-free guidance \cite{classifierfree} is employed by randomly replacing conditional tokens with zero tokens with a probability of 0.25.

%% file: sec/4_results.tex
\section{Results}
\label{sec: Results}

\subsection{Visual results} 
\label{sec: visual_Results}
We present text-conditioned generations in Fig. \ref{fig:results}. 
\modelname is capable of generating animatable and realistic 3D humans that accurately align with the input text descriptions. Our method supports a wide range of variations, including gender, age, as well as diverse clothing styles and hairstyles.
In Fig. \ref{fig:teaser}, we demonstrate sequential editing using text, scribble images, and body part images as inputs. 
Please refer to the supplemental material for details about the editing implementation and additional visual results.

%

\begin{table*}[t]
\centering
\caption{Quantitative comparison results with SOTA methods. \textcolor{1st}{$\blacksquare$} and \textcolor{2nd}{$\blacksquare$} denote the 1st and 2nd places. }
\scalebox{0.82}{ 
\begin{tabular}{@{}c|cc|cc|ccc|c@{}}
\toprule
\multirow{2}{*}{Method} & \multicolumn{2}{c|}{Semantic Align} & \multicolumn{2}{c|}{Quality}  & \multicolumn{3}{c|}{FID  $\downarrow$ } &\multirow{2}{*}{Inference Time$\downarrow$}    \\ 
 & BLIP-VQA $\uparrow$ & Text CLIP Score $\uparrow$  & CLIB-FIQA $\uparrow$  &HyperIQA $\uparrow$ & 2D Diffusion  & THuman 2.0 & HuGe100K  &\\
 
\midrule
TADA \cite{TADA}  &{0.50}  &{0.71}&{0.48} &{55.02}   &{188.19}   & {N/A}  & {N/A}  &{2.5h} \\ 
HumanGaussian \cite{HumanGaussian} &\cellcolor{2nd}{0.62}  &{0.73}&{0.39}&{33.61} &{239.33}    & {N/A}    & {N/A} &{1.2h} \\ 
 DreamWaltz-G \cite{DreamWaltz-G} &{0.58}  &\cellcolor{2nd}{0.75} &\cellcolor{2nd}{0.50}&\cellcolor{2nd}{59.33} &{175.23}  & {N/A}  & {N/A}&{3.0h}\\  
TeRA \cite{TeRA} &{0.42}  &{0.67}  &{0.44}  &{44.01}  &\cellcolor{1st}{151.80} & {N/A} & {N/A}&\cellcolor{2nd}{12s} \\ 
SIGMAN \cite{SIGMAN} &{0.29}  &{0.58}&{0.42} &{56.11} &{280.06} &  \cellcolor{2nd}{121.40}& \cellcolor{2nd}{160.48} & \cellcolor{1st}{3s}\\

  \modelname &\cellcolor{1st}{0.64}  &\cellcolor{1st}{0.76} &\cellcolor{1st}{0.55} &\cellcolor{1st}{63.05}  &\cellcolor{2nd}{160.91 } & \cellcolor{1st}{96.03} & \cellcolor{1st}{76.50} &\cellcolor{2nd}{12s} \\ 
   
\bottomrule 
\end{tabular}

\label{tab: Quantitative_Comparison}
} 
\end{table*}

\begin{table}[t]
\centering 
\caption{User studies. \textcolor{1st}{$\blacksquare$} and \textcolor{2nd}{$\blacksquare$} denote the 1st and 2nd places. }
\scalebox{0.65}{ 
\begin{tabular}{@{}c|cccc@{}}
\toprule
\multirow{2}{*}{Method} & \multicolumn{4}{c}{User study $\uparrow$ }                \\ 
 & {Semantic align.} & {Consistency} & {Visual quality}  &{Geometric quality}  \\  
\midrule
TADA \cite{TADA}      & {2.89}  & {3.18}  & {2.29}  & {2.15}\\ 
HumanGaussian \cite{HumanGaussian}  & {3.59}  & {3.37}  & {2.79}  & {2.77}\\ 
 DreamWaltz-G \cite{DreamWaltz-G} & \cellcolor{2nd}{3.68}  & \cellcolor{2nd}{3.93}  & \cellcolor{2nd}{3.41}  & \cellcolor{2nd}{3.37}\\ 
TeRA \cite{TeRA}  & {2.63}  & {3.74}  & {3.30}  & {3.28}\\ 
SIGMAN \cite{SIGMAN}    & {1.65}  & {1.86}  & {1.40}  & {1.52}\\ 
  \modelname   & \cellcolor{1st}{4.56}  & \cellcolor{1st}{4.68}  & \cellcolor{1st}{4.65}  & \cellcolor{1st}{4.63}\\ 
\bottomrule 
\end{tabular}
\label{tab: User study}
} 
\end{table}

\subsection{Comparison} 
\label{sec: Comparison}

%
We compare \modelname with SOTA SDS-based 3D full-body avatar generation methods (TADA \cite{TADA}, HumanGaussian \cite{HumanGaussian}, and DreamWaltz-G \cite{DreamWaltz-G}) and diffusion models that directly model the 3D human distribution (TeRA \cite{TeRA} and SIGMAN \cite{SIGMAN}).
Additionally, we provide comparisons with 3D human reconstruction methods in the supplemental material.

\subsubsection{Qualitative comparison}
\label{sec: Qualitative_Comparison}
Using the same text prompts, we show examples generated by SOTAs and \modelname in Fig.~\ref{fig: comparison}.
%
All SDS-based methods exhibit unrealistic visual styles. Both TeRA and SIGMAN demonstrate poor semantic alignment compared to other approaches. In the nurse case, although TeRA successfully generates a nurse avatar, it fails to produce the correct color of the uniform (``{pastel pink nurse’s uniform}''). SIGMAN fails to generate an aligned appearance in both cases. Additionally, TeRA suffers from a synthetic appearance, whereas SIGMAN demonstrates low visual quality. In contrast, \modelname produces superior generation results in both semantic alignment and color, with realistic facial details and overall higher fidelity.

\subsubsection{Quantitative comparison}  
\label{sec: Quantitative_Comparison}
We use 50 text prompts as inputs to generate 50 avatars for each method. 
Each avatar is rendered from multiple viewpoints (frontal, side, and back), resulting in three rendered images per avatar for evaluation.

\noindent\textbf{Semantic alignment.} 
We use BLIP-VQA from Progressive3D \cite{Progressive3D} to measure semantic alignment. Additionally, we estimate captions from the rendered images, and compute the CLIP feature distance between the estimated captions and the ground-truth text (Text CLIP score). 

\noindent\textbf{Visual quality.}
To evaluate the quality of the rendered avatar images, we employ CLIB-FIQA \cite{CLIB-FIQA}, a specialized method for assessing human facial image quality. For full-body avatar image quality assessment, we adopt HyperIQA \cite{HyperIQA}.
We report FID \cite{FID} between the renderings of text-generated avatars and 2D diffusion-generated images (using the same text).
Note that SDS methods and TeRA cannot be conditioned on images, making it infeasible to report metrics of them on large-scale image datasets.  
For methods that support image as inputs (SIGMAN and \modelname), we generate avatars using 200 images from THuman 2.0 \cite{Thuman} and 200 from HuGe100K \cite{IDOL}, and compute FID between the avatar renderings and the ground-truth images. 
Further details of evaluation metrics are provided in the supplemental material.

\noindent\textbf{User study.}
We recruited 30 participants to evaluate rotating videos of 50 text-generated avatars produced by different methods. Each participant was presented with the results of 10 randomly selected avatars and asked to rate them on a 5-point scale across four criteria: text alignment, multi-view consistency, visual quality, and geometric quality. The questionnaire template used in the user study is provided in the supplementary material.

Tabs. \ref{tab: Quantitative_Comparison} and \ref{tab: User study} summarize our quantitative evaluations. 
Our \modelname outperforms all state-of-the-art methods in semantic alignment, visual quality, and human preference by leveraging large-scale, real-world video data. 
In contrast, SDS-based approaches rely on 2D diffusion models to achieve strong semantic alignment, but this results in reduced visual quality. 
Meanwhile, 3D human diffusion model counterparts are trained on much smaller datasets, which negatively impacts both respects.
Regarding FID, TeRA is trained on diffusion-generated images and therefore naturally aligns with the 2D diffusion distribution. \modelname achieves improved FID on image datasets compared to SIGMAN.

 \begin{figure*}[t]
  \centering
  \includegraphics[width=0.99\linewidth]{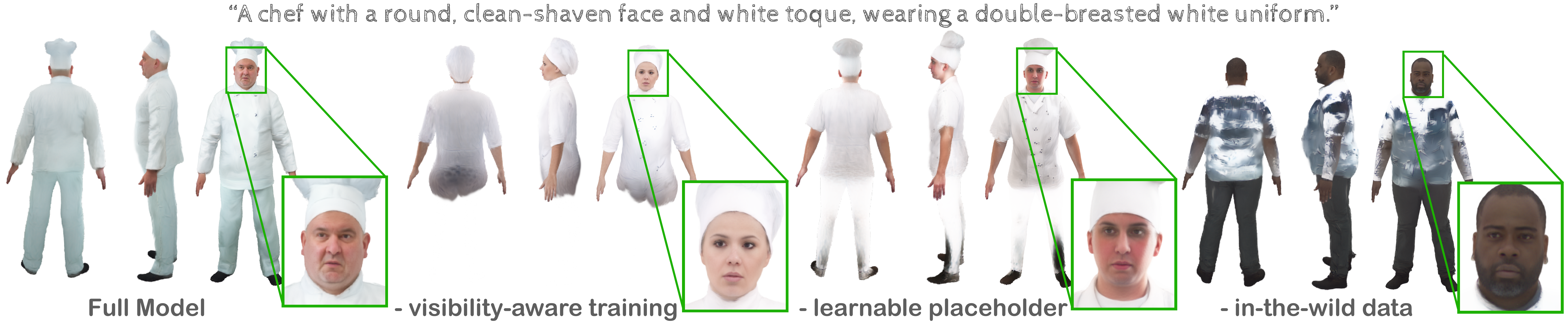}
  
  \caption{ 
  We conduct ablation studies by individually removing the visibility-aware training, learnable placeholder, and in-the-wild training data components to demonstrate the effectiveness of each.
   } 
  \label{fig:ablation}
\end{figure*}

\subsection{Ablation studies}
\label{sec: Ablation_studies}

We conduct ablation studies to evaluate the effectiveness of the proposed training strategies.
Fig.~\ref{fig:ablation} shows the comparison results. Additional evaluations are provided in the supplemental material.

\noindent\textbf{Visibility-aware training.} We assess the impact of the visibility-aware training strategy by training the diffusion model directly on all tokens. We include both valid and invalid tokens as training data, and perform loss computation on all tokens.
Without visibility-aware training, the model exhibits noticeable blurriness and transparency in the lower and back body, similar to the invalid regions present in the training data.

\noindent\textbf{Learnable placeholder.} To validate the effectiveness of the learnable placeholder, we replace all invalid region with fixed zero tokens during visibility-aware training.
We observe that in the absence of a learnable placeholder, the generated avatars display unnatural color.

\noindent\textbf{In-the-wild data.} To evaluate the generalizability provided by in-the-wild data, we train \modelname exclusively on indoor capture data, containing 3,000 identities.
Without in-the-wild data incorporated into the training set, the model overfits to the captured data and fails to generate text-aligned results, showcasing poor generalization.

%


 

%% file: sec/5_conclusion.tex
\section{Conclusion}
\label{sec: Conclusion} 
\modelname achieves state-of-the-art quality through data scalability enabled by large-scale, imperfect real-world videos. This effective utilization of imperfect data is realized by (i) using a feed-forward reconstruction model to tokenize real-world videos, and (ii) introducing a visibility-aware training scheme that handles partial observations. 
Experiments show that leveraging real-world videos significantly improves both the diversity and generalizability of the model, while the visibility-aware scheme filters unreliable signals to maximize use of high-quality data.
The resulting 3D avatar diffusion model charts a path toward 2D-scale training for 3D digital humans. 

The quality of \modelname is constrained by its reliance on Linear Blend Skinning inherited from the reconstruction model for animation, which can lead to unrealistic deformations, particularly for loose clothing under extreme poses (see examples in supplemental material).  
For future work, we aim to further strengthen the reconstruction model to boost fidelity and drivability, and to expand data scale for continued gains.
